\documentclass{article}
\usepackage{arxiv}
\usepackage[utf8]{inputenc} 
\usepackage[T1]{fontenc}    
\usepackage{hyperref}       
\usepackage{url}            
\usepackage{booktabs}       
\usepackage{amsfonts}       
\usepackage{nicefrac}       
\usepackage{microtype}      
\usepackage{lipsum}		
\usepackage{graphicx}
\usepackage{doi}
\usepackage{longtable}
\usepackage{caption}
\usepackage{subcaption}
\usepackage{multirow}
\usepackage[font=small,labelfont=bf]{caption}
\usepackage{array}

\usepackage[table]{xcolor}
\definecolor{degree1}{RGB}{70, 130, 180}    
\definecolor{degree2}{RGB}{255, 165, 0}     
\definecolor{degree3}{RGB}{220, 20, 60}     

\definecolor{degree1bg}{RGB}{70, 130, 180}    
\definecolor{degree2bg}{RGB}{255, 165, 0}     
\definecolor{degree3bg}{RGB}{220, 20, 60}     

\definecolor{degree1text}{RGB}{70, 130, 180}    
\definecolor{degree2text}{RGB}{255, 140, 0}     
\definecolor{degree3text}{RGB}{180, 20, 60}

\title{Consciousness in Artificial Intelligence? A Framework for Classifying Objections and Constraints}

\date{} 					

\author{%
  Andres Campero\thanks{Work done partly at MILA with a grant from Open Philanthropy. Correspondance: campero@mit.edu} \\
  AE Studio\\
  \And
   Derek Shiller \\
   Rethink Priorities\\
   \And 
   Jaan Aru\\
   Institute of Computer Science\\
   University of Tartu \\
   \And
   Jonathan Simon\thanks{Work supported by grants from Open Philanthropy, SSHRC and FRQSC.} \thanks{We thank David Chalmers for comments on an earlier draft.} \\
   Department of Philosophy\\
   Université de Montréal\\
}

\begin{document}
\maketitle

\vspace{10pt}
\setcounter{footnote}{0}
\begin{abstract}
We develop a taxonomical framework for classifying challenges to the possibility of consciousness in digital artificial intelligence systems. This framework allows us to identify the level of \emph{granularity} at which a given challenge is intended (the levels we propose correspond to Marr's levels) and to disambiguate its degree of \emph{force}: is it a challenge to computational functionalism that leaves the possibility of digital consciousness open (degree 1), a practical challenge to digital consciousness that suggests improbability without claiming impossibility (degree 2), or an argument claiming that digital consciousness is strictly impossible (degree 3)? We apply this framework to 14 prominent examples from the scientific and philosophical literature. Our aim is not to take a side in the debate, but to provide structure and a tool for disambiguating between challenges to computational functionalism and challenges to digital consciousness, as well as between different ways of parsing such challenges.

\end{abstract}


\vspace{10pt}

\section*{Introduction}

Might digital AI systems soon be conscious? Are some of them already? In a recent work, Butlin et al. \cite{butlin2023consciousness} presented some reasons for answering in the affirmative by identifying leading theories on which consciousness requires only certain kinds of computational organization that we see in contemporary AI systems. In this paper, we survey reasons for answering in the negative (along with possible replies), while disentangling the often confounded objections to the possibility of digital consciousness, from objections to the idea that consciousness is determined by computational organization.

By `consciousness', we mean phenomenal consciousness, or in the words of Nagel \cite{nagel1980like}, the fact of there being “something it is like” to be the system in question, i.e., whether the system has “qualia” or “subjective experience” – the form of consciousness about which there is a hard problem \cite{chalmers1995facing}. Paradigmatic examples in humans include the feeling of a scraped knee, the visual experience of color, and the taste of chocolate. What about `sentience'? Some in the literature use this term as a synonym of `consciousness', others as 
reserved for \emph{valenced} forms of consciousness like pleasure and pain. Either way sentience requires consciousness: our focus here is the latter.

By `computational functionalism', we mean the popular view according to which consciousness is determined by computational organization. As we will see, the exact definition of `computational organization' is not precisely established, but it roughly means ``physical processes in which information is processed in systematic ways''. The question of whether digital AI systems are now or soon will be conscious is closely tied to the question of whether computational functionalism is true. Many who accept that current or near-future AI systems may be conscious embrace computational functionalism in some form (see \cite{butlin2023consciousness, campero2024report} for surveys). Further, some of the more dramatic challenges to computational functionalism challenge the possibility that anything running on digital hardware could be conscious. 

However, there are good reasons to distinguish the question of whether computational functionalism is true from the question of whether digital AI systems are now or soon will be conscious. First, quite simply, the answers to the two questions might come apart. It is possible to be a computational functionalist but still categorically deny that digital AI systems (e.g., those running on von Neumann or SIMD/SIMT architectures) can be conscious, e.g. if you think the computation in question must be analog, or hybrid \cite{piccinini2013neural}. Conversely, one might reject computational functionalism but still accept that digital AI systems can be conscious (as long as their hardware fulfills certain architectural conditions that do not quite count as `computational'). 

A second reason to distinguish the questions is that `computational functionalism’ can be understood in different ways. Framed in terms of Marr’s \cite{Marr1982} levels, it might be interpreted as either a claim at the \emph{computational} or at the \emph{algorithmic} level (more on this below). There are important interpretive questions here, but we don’t have to settle them in order to assess whether specific contemporary digital AI systems can be conscious. Our framework allows us to bypass these thorny interpretive questions for the purpose of classifying reasons to doubt the possibility of digital consciousness.


We emphasize that our aim in this survey is not to muster a new ``strength in numbers'' argument against digital consciousness, by highlighting how many different routes there are to that conclusion. We, the authors of this piece, disagree over how to appraise the various arguments we survey. Instead, our primary aim is understanding, articulation and disambiguation, rather than persuasion. 

For those readers who take digital consciousness to be a serious possibility, we envision this paper may help to structure the debate and clarify the points of disagreement that a defense of digital consciousness may wish to address. 

For those who are more skeptical, this paper should help put the reasons for skepticism into context. This is not as trivial as it might seem, because these reasons can be quite diverse. For one thing, as we will see, some of them are generally quite compatible with thinking of the mind as computational in a broad sense, while others are certainly not. For another thing, we find that many ideas in the literature that seemingly designate a single line of thinking, in fact name a range of different objections, with different presuppositions. “Enactivism”, for example, names an entire family of objections, at different levels of abstraction and generality. \footnote{“Enactivist” objections to digital consciousness might be construed at various levels. At the level of input-output mappings, as the claim that consciousness involves non-computable dynamical coupling \cite{landgrebe2022machines}. At the level of algorithmic organization, as the claim that consciousness involves analog computation that can only be emulated, but not implemented, on digital systems. At the level of “implementation” as the claim that biological complexity of the implementing system is necessary for consciousness \cite{aru2023feasibility, varela1991embodied, thompson2010mind}, or the claim that consciousness requires being a part of a living system, irrespective of what that means about one’s organization.  A taxonomy like ours allows us to not only map out existing debates but further refine and disambiguate them.} 

Distinguishing these arguments allows us to better assess them in turn. Conversely, in some cases seemingly distinct objections in fact hinge on common argument patterns, at least under certain disambiguations. For example, Godfrey-Smith’s fine-grained functionalism \cite{godfrey2016mind} has much in common with Thompson’s version of enactivism \cite{thompson2010mind}.

Crucially, we do not suggest that it is possible to settle this issue once and for all: the epistemically cautious approach here is to allocate credences and assess precisely how confident in each proposition we ought to be. The stakes are high: whether digital systems can be conscious can dramatically alter the normative landscape. This means that as long as we find that we have any credence left over, after allocating our credence to the various objections we take to populate the taxonomy outlined here (note again that we do not provide an exhaustive catalogue here but only a framework for doing so) the prospect of digital consciousness will still merit attention \cite{long2024taking, bengio2025illusions}.

Our taxonomy draws two key distinctions: level of granularity, and degree of force. Our proposed levels of granularity are, in essence, Marr's levels:

\begin{itemize}
\item Level 1: objections and constraints at the level of input/output mappings
\item Level 2: objections and constraints at the level of algorithmic organization
\item Level 3: objections and constraints at the level of underlying physical structure and implementation

\end{itemize}

Our proposed scale of degrees of force is as follows:

\begin{itemize}
\item Degree 1: Objections against computational functionalism , according to which consciousness depends on something else, instead of or in addition to computational structure (but that do not directly rule out consciousness in a digital system)
\item Degree 2: Objections according to which consciousness may depend on computational structure, but there are significant challenges to realizing the right structure in a digital system
\item Degree 3: Objections according to which consciousness may depend on structure, but it is impossible to realize the right structure in a digital system
\end{itemize}




The rest of the paper is organized as follows. The next section elaborates on each of the levels of abstraction, explaining how we use Marr’s hierarchy and the relevant notions of computation, algorithm and implementation; and discussing a critical ambiguity between two conceptions of computational functionalism, one (Turing’s approach) which is best understood as a claim at the “computational” level and the other (endorsed by Chalmers, Piccinini, Sprevak, Klein and others) which is best understood as a claim at the “algorithmic” level. The following three sections present the objections and constraints organized at each level of abstraction. The final section concludes.

\begin{table}[t]
\centering
\small
\renewcommand{\arraystretch}{1.3}
\begin{tabular}{p{2.1cm} p{5.2cm} p{7.8cm}}
\toprule
\multicolumn{3}{c}{\large\textbf{Objections and Constraints to Digital Consciousness}} \\

\multicolumn{3}{c}{\rule{0pt}{3ex}\small\textcolor{degree1!60}{\textbf{Degree 1}} - \textbf{Orthogonal to Computationalism} \quad \textcolor{degree2!60}{\textbf{Degree 2}} - \textbf{Significant Challenges} \quad \textcolor{degree3!60}{\textbf{Degree 3}} - \textbf{Precludes Consciousness}\rule{0pt}{2ex}} \\
\midrule

\multicolumn{1}{c}{\textbf{Level}} & \multicolumn{1}{c}{\textbf{Objection}} & \multicolumn{1}{c}{\textbf{Description}} \\
\midrule

\multirow{4}{2.2cm}{\textbf{Level 1:} Input-Output Mappings} 
& \cellcolor{degree3!25}1. Beyond Turing Computability & Some non-computable ability is required for consciousness \\
& \cellcolor{degree3!25}2. Dynamic Coupling & Environment-system coupling cannot be approximated \\
& \cellcolor{degree2!25}3. Computational Complexity & Complexity of consciousness is practically unimplementable \\
\midrule

\multirow{4}{2.2cm}{\textbf{Level 2:} Algorithmic Organization} 
& \cellcolor{degree1!25}4. Engineering-Architectural Dimensions & Architectural constraints place limits on implementation \\
& \cellcolor{degree1!25}5. Physical Time & Physical time constrains conscious experience \\
& \cellcolor{degree3!25}6. Analog Processing & Analog processing is necessary for consciousness \\
& \cellcolor{degree2!25}7. Notion of Representation & Conscious representation requires downstream interpretation \\
\midrule

\multirow{5}{2.2cm}{\textbf{Level 3:}\\ Physical Organization and Implementation} 
& \cellcolor{degree1!25}8. Triviality and Counterfactual Problems & Computation is counterfactually loaded unlike consciousness \\
& \cellcolor{degree3!25}9. Causal Structure (IIT) & Causal power determines consciousness \\
& \cellcolor{degree3!25}10. Slicing Problems & Multiple realizability implies violation of unity of experience \\
& \cellcolor{degree3!25}11. Electromagnetic Fields & EMF topology constrains consciousness  \\
& \cellcolor{degree2!25}12. Biological Complexity & Inextricable integration of subcellular, neural and system levels \\
& \cellcolor{degree1!25}13. Biology is Fundamental & Self-organization and enactment are fundamental \\
& \cellcolor{degree1!25}14. Quantum Properties & Quantum aspects determine consciousness \\

\bottomrule
\end{tabular}
\vspace{.1cm}
\caption{Summary of objections by abstraction level and degree of challenge: 1) consciousness is orthogonal to computational functionalism, 2) digital consciousness faces significant practical challenges, or 3) consciousness is impossible in digital machines.}
\label{table:taxonomy}
\end{table}

\section*{The Taxonomy}

Marr \cite{Marr1982} distinguishes three levels at which information processing systems may be understood: the computational, algorithmic and implementational levels. We will elaborate on how we understand this scheme, and then use it to taxonomize objections to the prospect that a digital implementation of an information processing system is conscious. 

\subsection*{	1) Objections and constraints at the level of input/output mappings}

At the computational level, one classifies systems in terms of the problem to be solved, i.e. the function (understood as a set of {input, output} pairs) that characterizes the system's behaviour. In 1936, Alonzo Church and Alan Turing proposed formal characterizations of what it means for a function to be solvable by a method. Turing, in an appendix to his 1936 paper \cite{turing1936computable}, proved the extensional equivalence of his and Church’s two formal characterizations. Remarkably, other approaches, such as von Neumann’s, also proved to be equivalent to these two\footnote{These equivalences further led to the philosophical claim (the Church-Turing Thesis) that the class of functions singled out by Church, Turing and von Neumann captures the extent of any intuitive conception of any method for calculation or computation \cite{sep-church-turing}}. Since then, as established by the field of Theory of Computation, this equivalence to the set of functions solvable by Turing’s method (a Turing Machine) has become the most widespread definition of computability \cite{sipser1996introduction}.

Here we arrive at the first point where a thesis deserving the title 'computationalism' might be defended: one might argue that as a necessary condition, there are some computable functions that a mind must be able to solve. One might also go even further and suggest that this is not only necessary but sufficient: that there is some class of computable functions such that all it is to be a mind is to be able to compute them. Consider Turing’s own suggestion that displaying the linguistic behavior required to pass the Turing test suffices for intelligence \cite{turing1950computing}.
Note, crucially, that the equivalence discussed in these foundational works is extensional rather than intensional. While Turing’s appendix shows how to convert an effective method from lambda calculus into an effective method for a Turing Machine (and vice versa), this does not mean that the two methods are identical, but only that they map the same inputs to the same outputs. Generally, multiple effective methods can implement a function. For example, one can use bubble sort or merge sort to derive a sorted array (and there can be clear reasons to use one rather than the other). This highlights the importance of distinguishing between the question of which functions a system can compute and the question of which effective methods (i.e. algorithms) it uses to do so.

This is a common source of confusion. For our purposes, objections to the possibility of digital consciousness that live at level 1 are, in effect, objections to the idea that digital systems exhibit the right kind of input/output mappings for consciousness. Namely, to endorse such an objection, one needs to hold that there are some problem-solving capacities essential for consciousness that digital systems lack (e.g., the capacity to “solve” non-computable functions).

\subsection*{2) Objections and constraints at the level of algorithmic organization}

This brings us to Marr’s algorithmic level. There are a range of things one might mean by 'algorithm'. As we noted, the equivalence between the Turing and Church (and later, von Neumann \cite{vonneumann1945first}) formulations is not an equivalence of effective methods, but rather of the class of functions solvable by each formalism. Namely, it is not the case that an effective method using lambda calculus for solving a given function is the same as an effective method using a Turing machine; rather, one can emulate (or virtually implement) one kind of system within the other. Further, as with the example of bubble sort and merge sort, even on a Turing machine, there will in general be distinct effective methods for solving the same function, just as there are often distinct proofs of the same theorem. 

One might well suppose that consciousness is sensitive not only to which functions one solves for, but also how one solves for them – which effective method one uses. Crucially, this is not necessarily to reject computational functionalism but rather to refine what one means by it.

To establish precisely what makes two physical processes implement the same effective method is subject to debate and requires further philosophical and computer-theoretical work. Notoriously, Putnam, Searle and others \cite{putnam1988representation, searle1980minds} argue that there is no intrinsic, non-trivial way of answering this question: in their view, the individuation of effective methods in physical systems is in the eye of the beholder. Others such as Chalmers and Piccinini \cite{chalmers2011computational, piccinini2013neural} resist, arguing that there are causal, counterfactual, or mechanistic conceptions of the relevant notion; so that for example, a sequence of state transitions formalized as finite state automata (FSA) characterizes an effective method (which helps establish the idea that bubble sort on a Mac is the same type of effective method than bubble sort on a Lenovo).

We will consider subtleties deriving from this debate below because some of the objections we consider hinge on them. Crucially, this level is home to a number of interesting objections to digital consciousness, and many of them are consistent with the letter or with the spirit of computational functionalism, depending on how precisely we understand that view.

At this level, for example, we find one version of the objection that only analog systems can be conscious (because, though the processes in the brain that realize a mind are computational processes, their analogicity is essential). We also find objections that timing, parallelism or synchronicity constraints on a computational architecture might matter for consciousness. Because of the way that algorithms are usually defined as pure state transitions in computational theory, taking architecture to be relevant in this way would violate the letter of computational functionalism, though arguably not the spirit, as enriched conceptions of computational process that leave room for fine-grained architectural distinctions are all that some of these constraints call for.

\subsection*{3) Objections and constraints at the level of underlying physical structure and implementation}

We now turn to the third level, objections to digital consciousness that more definitely reject both the spirit and the letter of computational functionalism. 

Computational organization does not encompass all organization. In particular, it is an abstract form of organization that implies some degree of medium independence, multiple realizability, and independence across layers of abstraction (how much depends on how we construe the implementation relation). But physical systems can have causal, structural or organizational properties that are not multiply realizable or independent in the relevant sense. As we will see, if these properties are relevant, they go beyond computation.

In the case of the brain in particular, some argue that too much credence has been given to the so-called Neuron Doctrine, which takes the activity of action potentials to be the only aspect of brain activity necessary to explain cognition (and consciousness) \cite{aru2025assumptions}. According to critics, this underrepresents the real complexity of the brain’s activity that is relevant for consciousness.  

This is another example of a line of criticism that can be interpreted in different ways, corresponding to the levels and degrees of our taxonomy. Read as a constraint at level 3 (i.e., concerning underlying physical structure and implementation) it might suggest that the underlying neurochemistry or causal structure matters for consciousness. Read at level 2, in contrast, it might suggest analog constraints on the underlying process. Read at level 1, it might suggest that the relevant computation is too complex for any digital system to solve in polynomial time. Alternatively, we might read opposition to the Neuron Doctrine as not an objection to digital consciousness at all, but merely a constraint on which algorithms might realize it.


We observe that in principle there are also objections at the implementation level that do not hinge on structure at all, but instead on \emph{intrinsic natures}. Such an objection might claim that, even were a perfect silicon duplicate of a human being possible (which is not obvious : if such a being lived in an oxygen-rich environment it would exhale silicon dioxide, i.e. sand) it would not be conscious, because consciousness depends on the intrinsic nature of carbon. One might also mention quantum properties here.  However, though one sometimes sees the ``biological view'' summarized in this way, this line of thinking fits more naturally with panpsychism or panprotopsychism: proponents of the biological view may be more likely to say that carbon simply is a better material to build with, since it has has stronger bonds and its oxidation products are gases. 


\subsection*{Other Arguments Against Digital Consciousness}

Before proceeding, we emphasize that our aim is to provide and illustrate a toolkit for disambiguation, rather than to comprehensively catalogue every possible objection. We also acknowledge that there is room for disagreement in some cases over where in the taxonomy, precisely, a given view should go. This is unavoidable: one of the key lessons we hope you take away from the paper is that it is essential in the context of a specific discussion to clarify the level(s) and degree(s) at which you understand the argument you are discussing, but this can vary from one discussion to another. 

For this reason, there are several important arguments in the literature (about the nature of consciousness) that we do not attempt to classify. A first case in point: the arguments in the reductionism vs emergentism debates (e.g. Jackson's Mary argument  \cite{jackson1982epiphenomenal} and Chalmers' Conceivability and Scrutability arguments \cite{Chalmers2012-CHACTW}). Of course if dualism is true then there is something ``more'' to consciousness than computational organization, metaphysically speaking (which might suggest a degree 1 style constraint). But the right computational organization still might suffice, as a matter of psychophysical law, for consciousness, if dualism is true, as Chalmers \cite{chalmers1996absent} has argued. Alternatively, we might envision a variant of panpsychism on which only things made of carbon have the right intrinsic profile for consciousness, which gives you a level 3 degree 3 objection.

Another example: arguments that specific current algorithms are missing something crucial to achieve general intelligence or understanding, including those developed by AI researchers such as Josh Tenenbaum \cite{lake2017building}, Gary Marcus \cite{marcus2018deep}, Emily Bender and Alexander Koller \cite{bender-koller-2020-climbing}, and Yann Lecun \cite{lecun2022path}. Here, too, there are different ways of filling in further details that would lead to constraints at different locations in our taxonomy. If you take these arguments to be evidence for a pessimistic induction that every digital computational approach will have similar limitations to current approaches, then you have a level 2 degree 2 objection on your hands. If, alternatively, you take these arguments as pointing the way toward some other computational approach that will work (e.g. one requiring a neurosymbolic hybrid, or energy-based modelling), then you may construe these arguments as strictly orthogonal to the question of digital consciousness, or indeed in support of it.

We observe that some of the most famous arguments against computational functionalism also require further interpretation and context, in order to be parsed into specific challenges to digital consciousness. Consider Block's Nation of China argument \cite{block1980troubles}. Is the moral that any system engineered for computing cannot be conscious, or just that there are some architectural constraints on the spacing and timing of computational processes? Or consider Searle's Chinese Room argument \cite{searle1980minds}. Is it targeted at specific algorithms (viz., classical symbolic ones, drawing on the analogy between a CPU and a human secretary working punchcards) or is it meant to refute any sort of computational intelligence (and by extension, arguably, consciousness), even for `subsymbolic' algorithms where the apt analogy for a processing unit is not a human secretary but a stream of neurotransmitters?

To wrap up: there are many important arguments in the literature on the nature of consciousness that are relevant to the possibility of digital consciousness, but where to say exactly how we must fill in further details. In a few cases below we do this : for example we will identify versions of ``enactivist'' challenges to digital consciousness living in various regions of our taxonomy. But our general aim is not to populate the taxonomy, it is to illustrate how asking the question, ``at what level and degree do you understand this objection?'' can be helpful as we attempt to settle our views on the prospects for digital consciousness. Doing so helps us map out the logical space, clarifying which premises a given argument relies on and how these are supported.

\pagebreak

\section*{Level 1, Objections and Constraints on Input-Output Mappings} 

Objections to the possibility of digital consciousness at this level directly challenge computational functionalism, construed as the thesis that some set of computable input-output mappings suffice for consciousness, by suggesting that consciousness goes beyond the capabilities of a Turing Machine, and hence beyond the capabilities of digital computers.

\subsection*{\underline{1. Consciousness Is Non-Computable for Gödelian Reasons (Level 1, Degree 3) }}

John Lucas and Roger Penrose\cite{penrose1989emperor} have developed the idea, drawing on Gödel's Second Incompleteness Theorem, that human intellect encompasses truths that no formal or computational system can derive. In brief, Gòdel shows that no sufficiently expressive formal system can prove its own consistency, but insightful humans can nevertheless intuit the consistency of many such formal systems. Thus for most any given (consistent) formal system we can know something that the system cannot; viz., that it is consistent.


\subsection*{\underline{2. Consciousness is Non-Computable for Reasons from Chaos and Dynamical Systems Theory (Level 1, Degree 3)}}
Landgrebe and Smith \cite{landgrebe2022machines} argue that consciousness depends on chaotic dynamical coupling between brain, body and environment, where in contrast digital computational systems are engineered to suppress chaotic effects. Moreover, digital computational systems can at best numerically approximate solutions to the differential equations that describe complex, chaotic dynamical systems.

Indeed, Pour-El and Richards have shown that in some cases, exact solutions to differential equations for dynamical systems are not only intractable but strictly non computable by discrete systems \cite{pour1983computability} (see Simon \cite{simon2024intelligence} for discussion).

\subsection*{\underline{3. Consciousness is Computationally Intractable (Level 1, Degree 2)}}

The previous objections contend that consciousness is strictly non-computable. One might alternatively argue that consciousness, though computable \emph{sensu stricto}, is of too high of a computational complexity class to be tractable, i.e., that if run on digital hardware consciousness would amount to a super-exponential time algorithm.\cite{walsh2017singularity, landgrebe2022machines}.

\section*{Level 2, Objections and Constraints on Algorithmic Organization}

Objections to the possibility of digital consciousness at this level are concerned with the character of algorithmic processes. Even though when we speak of algorithms we envision specific processes or recipes for getting from input to output, still we abstract from many details, including details of architecture that software engineers must take into account in practice, such as timing constraints. Our ``degree one'' classification fits many of these : in general these go against the letter of (digital-compatible) computational functionalism, but not against its spirit (in the sense that they do not significantly lessen the likelihood that digital systems will be conscious).  However, some objections at level 2 do target the possibility of digital consciousness, such as objections that insist that the algorithms relevant to consciousness are analog. Other objections suggest that algorithm structure is not germane to consciousness one way or the other (because it is an interpretive question what algorithm something implements), undercutting many possible lines of defense of the prospect of digital consciousness.


\subsection*{\underline{4. Engineering-Architecture Dimensions (Level 2, Degree 1)}}
Algorithms are intended to be architecture-independent. Nevertheless, architectural considerations could be important for consciousness as they place constraints on the time and space it would take to implement the algorithm. We present three such considerations.

\textit{\textbf{Continuity and Interaction:}} The traditional definition of algorithms historically followed Turing’s paradigm which explicitly excluded the notion of interaction \cite{goldin2005church}. Nevertheless, the entanglement between inputs and outputs where later inputs to computation might depend on earlier outputs can be important \cite{petters2024we}. This objection goes more against the theoretical definition of algorithm than against practical computational software, such as Operating Systems, or AI models, which like the mind, have an interactive nature that continue to receive inputs mid-computation.

\textit{\textbf{Concurrency and Parallelism:}} Some architectures support true parallelism (paradigmatically for our purposes, that is the case for contemporary implementations of AI models and for the Brain), which is not directly modeled by FSAs or Turing Machines. While serial and parallel machines can run the same functions, they cannot use the same methods for computing those functions. In fact, from a theoretical perspective,  Petters and Jung argue that it has proven extremely difficult to pin down what a network of distributed computation actually computes and to compare the computational power of different approaches to distributed computation \cite{petters2024we, ritchie2023computing}.

\textit{\textbf{Control Flow:}} Programs must be compiled or interpreted before they are processed in a particular architecture. Borrowing the example from \cite{sprevak2007chinese}, “consider the difference between the programs that can be run on a Von Neumann-style PC and those that can be run on a computer based on Church’s $\lambda$-calculus. Imagine trying to run a PC version of Microsoft Word on a machine with a $\lambda$-calculus architecture. Such a program could not be run as it currently stands. A $\lambda$-calculus version of Microsoft Word would have to work in a radically different way. For one thing, on a $\lambda$-calculus architecture there are no loops, instructions, or stored variables, so the standard Von Neuman construction of a looped assignment statement could not be run.” While virtual machines are automated procedures for turning one type of program into another, the translation results in programs that can be significantly different, varying in the order in which instructions are executed, steps inserted and removed, etc. Constraints on control flow would go against the letter of some formulations of computational functionalism, but, we contend, would not go against its spirit in any deep sense, nor threaten the possibility of digital consciousness, unless the constraints proved especially difficult to satisfy.

\subsection*{\underline{5. Physical Time matters (Level 2, Degree 1)}}
Physical time does not have any role in models of digital computation (or in AI algorithms), which is individuated in terms of computational steps, independently of how much time it takes to compute a step. As Shiller observes \cite{shiller2024digital}, this opens difficult questions for a computational functionalist: “What happens if we pause a computation midway through an experience and leave it paused for 1,000 years? Does the experience stretch over that time or does the subject feel nothing over that interval? What if our computational system has a millennium-long clock speed such that it sits motionless for centuries – would its experiences be periodic?”. As O’Ritchie and Klein argue \cite{ritchie2023computing}, in an improved version of real-time computing, correct timing has to be constitutive of the computing task. This violates medium independence, and the letter of some definitions of computational functionalism -- but it points to an enriched conception of computational process (and accordingly of computational functionalism) rather than a rejection of the spirit of the view.\footnote{One can also ask the question about the difference between a dynamic algorithm described on the wall (in a purely spatial dimension), and a computer running the algorithm. While both preserve all functional relationships pertaining to the order of state transitions, architectural differences including temporal ones seem to make these different computational processes in an important sense.}

\subsection*{\underline{6. Analog Processing is Necessary (Level 2, Degree 3)}}
By design, digital computers are engineered to ignore their physical underlying analog properties. Brains, in contrast, rely on analog processing (whether it is the geometrical location of the synapses, the exact timing of an action potential, or the frequency of an electromagnetic field). Indeed, current evidence suggests that the functionally relevant aspects of neural processes depend on analog aspects. For example, temporal coding via spike timing may be critically important beyond spike rates in certain cases, such as smell \cite{piccinini2013neural}. 

The key question for us is whether, as far as consciousness is concerned, the brain’s analog methods are simply the means by which it implements some function which can also be implemented digitally, or whether it is analogicity per se that is necessary for consciousness. To take the former view is not necessarily to hold that all of the brain’s analog functions merely serve to approximate memory registers (or neural networks for which the Neuron Doctrine is true). Instead, we might think of the organizational requirements as calling for a (possibly digital) approximation of analogicity.\footnote{Maley emphasizes that the difference between analog and digital is not so much the continuous vs discrete nature, but rather the type of representation, where analog refers to representations that have an analogy to what they represent, in that magnitude changes of the representation reflect the magnitude changes of what they represent: “Analog representation is a kind of first-order representation (i.e. the representation of the magnitude of a number by physical magnitudes), whereas digital representation is a kind of second-order representation (i.e. the numerical representation of a number by its digits, which themselves are individually represented by variations in the values of a physical property)”. Digital representation abstracts away from its physical implementation, whereas analog representation takes advantage of it essentially without an intermediate medium independent representational level \cite{maley2021physicality}.}

However, it is known that there are things one can do with an analog device that one cannot do with a digital computer. Even though any continuous function can be approximated to arbitrary precision by a neural network (the Universal Approximation Theorem), there is no known way to approximate the solutions to an arbitrary set of differential equations using only a digital neural network to arbitrary precision (or numerical methods more generally), and indeed Pour-El and Richards have shown some differential equations are non computable by discrete systems but can be computed on analog computers \cite{pour1983computability, dyson2002life}. It remains to be seen whether our brains manage to compute anything that is not computable on digital computers, and it remains to be seen whether, if they do, any of these things are necessary for consciousness. If so, however, it would mean that digital consciousness is impossible (a fortiori as also would be any form of consciousness-preserving mind-uploading, see Piccinini \cite{piccinini2021myth} and Mandelbaum \cite{mandelbaum2022everything}). 

A couple of notes are in order. First, it is worth mentioning that even if this shows that strictly digital AI consciousness is impossible, still it does not show that neuromorphic AI is impossible. Our scope here is the question of digital AI consciousness, but the broader question of AI consciousness on near future hardware is of course closely related. Second, notice that the Landgrebe-Smith argument (an objection at level one) is that this intractability of dynamical systems means that consciousness is not computable. Here, we consider a different deployment of broadly the same considerations, to argue that while consciousness may be computable, the computations involved are essentially analog. This will not be the last example we consider in which different authors have deployed similar considerations in different ways.

\subsection*{\underline{7. Problematization of the Notion of Representation (Level 2, Degree 2) }}
AI researchers and computational neuroscientists alike use the notion of representation ubiquitously. The same goes for many computational theories of consciousness \cite{campero2024report, butlin2023consciousness}. The problem is that it is far from well established what counts as a representation, nor is it clear what is its role in computational functionalism. 

There is a rich literature on this topic (see \cite{sep-mental-representation} for a philosophical discussion, and \cite{shagrir2022nature} for a computational discussion. Most (but not all) agree that “mental states” in the brain are neither a manner of speech nor are derived from other’s interpretative stance. The question is whether the same is true for digital algorithmic systems: do the facts about what they represent depend on the facts about what we interpret them as representing? Some say yes -- this is the semantic view of computation.

In recent work, Rosa Cao argues that representations depend necessarilly on usage by some downstream user for some pragmatic purpose \cite{cao2022putting}. Relatedly, under the semantic (representational) view of computation, logical gates operate on physical states but also on truth conditions and are individuated by the ‘logical function’ they perform (AND, OR, NAND, …) which cannot be individuated by functional properties alone, for example Table \ref{tab:gate} recreated from \cite{sprevak2010computation, lee2021mechanisms} shows a device that is simultaneously implementing the operation AND (0V = 0; 5V = 1), and the operation OR (by relabelling to 0V=1, 5V=0). 

\begin{table}[h]
\caption{\textbf{AND or OR gate?} 
Which gate this simple voltage physical system is implementing depends on how the downstream user interprets the voltages. AND if (0V=0; 5V=1), OR if (0V=1; 5V=0)}
\label{tab:gate}
\vspace*{.5cm}
\centering
\setlength{\tabcolsep}{24pt}
\begin{tabular}{@{}ccc@{}}
\toprule
Input $a$ & Input $b$ & Output \\
\midrule
0V & 0V & 0V \\ 
0V & 5V & 0V \\  
5V & 0V & 0V \\ 
5V & 5V & 5V \\   
\bottomrule
\end{tabular}
\end{table}

Whether it is implementing OR or AND depends on the downstream user of the logical gate. One reading of this literature can result in a constraint on consciousness theories which want to both: i) rely on the notion of representation, while ii) not rely on the need of downstream representation-users. These two conditions are prevalent given the seemingly subjective individual nature of consciousness, and the ubiquity of the notion of representation in contemporary AI (The concept of learning representations is central to deep learning and current AI models \cite{bengio2017deep}), with one of the main conferences in the field even being called International Conference on Learning Representations (ICLR).

\section*{Level 3, Objections and Constraints on Physical Organization and Implementation}

Objections at this level challenge the fundamental premises of computational functionalism, violating or questioning some aspect of multiple realizability and medium independence, in some cases suggesting that the substrate of consciousness must be biological, or fundamental.

\subsection*{\underline{8. Counterfactual Structure and Triviality Problems (Level 3, Degree 1)}}
The core of this problem as stated by Klein \cite{klein2018computation} borrowing from \cite{maudlin1989computation} is that computation is a counterfactually loaded concept (it depends on what could have happened), while consciousness presumably only depends on what does happen.

This remark is related to the core of the so-called triviality problems to computation. The problem, as introduced by Putnam \cite{putnam1988representation} and discussed by Chalmers \cite{chalmers2011computational}, is that you can almost always find a mapping from physical states and transitions of any system to internal states and transitions of an FSA. 
So, for example, it has been argued that according to computationalism, every rock can implement every FSA. Notice that one can make an arbitrary mapping between the positions of the atoms in a rock (or any other physical process, such as the geometrical shape of the shadow of a leaf falling), and the sequence of state transitions in any computational process. To be sure, indeed if one knows how to make that mapping, one could (in principle, if not in practice) connect a monitor to the state of a rock and render the correct output (say an image represented as a bunch of states in the RGB color code).

A natural reply is that for some physical process to count as a computational implementation, what is required is not only a mapping for a particular computational run, but rather that a robust counterfactual structure expressed by conditionals must be satisfied. Namely, for a physical system to implement multiplication, it doesn’t suffice to map 3x3=9 for one run; what is required is the full mapping for all inputs and for multiple runs, time after time (understood as constrains on what outputs would have been generated, had other inputs been generated). It is only in virtue of this counterfactual reliability that computers are generally useful tools.  

This quickly gets into a rich and subtle philosophical discussion about the nature of computation. These include Putnam’s arguments that even the full counterfactual structure can be mapped (the trick is that one is free to partition and define states arbitrarily, including disjunctions (see \cite{chalmers1996rock, godfreysmith2009triviality})); as well as several replies and counterarguments to this and Maudlin’s arguments \cite{klein2008dispositional, bartlett2012computational, maudlin1989computation}.

Even if we assume this can somehow be settled and grant, like in Klein’s analogy \cite{klein2018computation}, that a videotape of my life can go through the same sequence of states without realizing a computation, let alone a mind, a version of the problem prevails: computation is counterfactually loaded, while consciousness seems not to be. Indeed two things can differ in computational status, without differing in what they are actually doing (as shown figure 2 with the example of a system which can be interpreted to implement the logical steps AND or OR depending on the downstream user). 

What then about consciousness? The objection here argues that consciousness should supervene on actual activity of the brain, not on what it could have been or have done in other circumstances\cite{gidon2025does}.\footnote{Notice this intuition is not universal. As we will see, the Integrated Information Theory for example is committed to the importance of the full cause-effect system, where even inactive elements count.} Indeed as pointed by Klein, anesthesia presumably works by blocking mechanisms that would have activated otherwise (e.g. affecting molecular effects or neurons that would have fired, not others that wouldn’t have fired anyway). At the very least, as discussed by Fekete and Edelman \cite{fekete2011towards}, if phenomenal experience depends on potentiality, an account of how that happens would seem to be required to provide a satisfactory explanation.\footnote{Notice that this can possibly be done, as a suggestive example, a value of 0 in a vector embedding of a machine learning model, indeed, carries information. So does a “non-firing” artificial neuron.}

\subsection*{\underline{9. Causal Structure : Integrated Information Theory (Level 3, Degree 3) }}
One of the most well established theories of consciousness is Integrated Information Theory (IIT) \cite{albantakis2023integrated} which starts with some axioms and provides a mathematical formalism that provides a metric ($\Phi$) of consciousness for every system based on the interconnectedness of its causal structure. In particular, causal power is different to functional organization, and depends on the physical connections independently of what can be virtually run at a higher level.

Some proponents of IIT claim that it is incompatible with both digital consciousness and computational functionalism, 
although understanding why requires some examination. In a recent tweet (\hyperlink{https://x.com/davidchalmers42/status/1813223307061199137}{https://x.com/davidchalmers42/status/1813223307061199137}) David Chalmers asks for an explanation of why Von Neumann architectures cannot be not conscious according to IIT. The question is the following: according to IIT the causal structure of a system is specified in terms of Probabilistic Automata (PA), some of which have non-zero $\Phi$. Von Neumann architectures can implement these PAs (in fact as stated by Chalmers, PAs are paradigmatic computational systems). Since IIT asks you to look at all the possible abstractions and subsystems keeping those that have the biggest $\Phi$, then why do they not also have a non-zero $\Phi$, and hence are conscious? The answer is that $\Phi$ is low for all the levels of abstraction, since even the virtual implementation is causally mediated by underlying components (such as the serial CPU). 

\begin{figure}[t]
\begin{center}
\includegraphics[width=\textwidth]{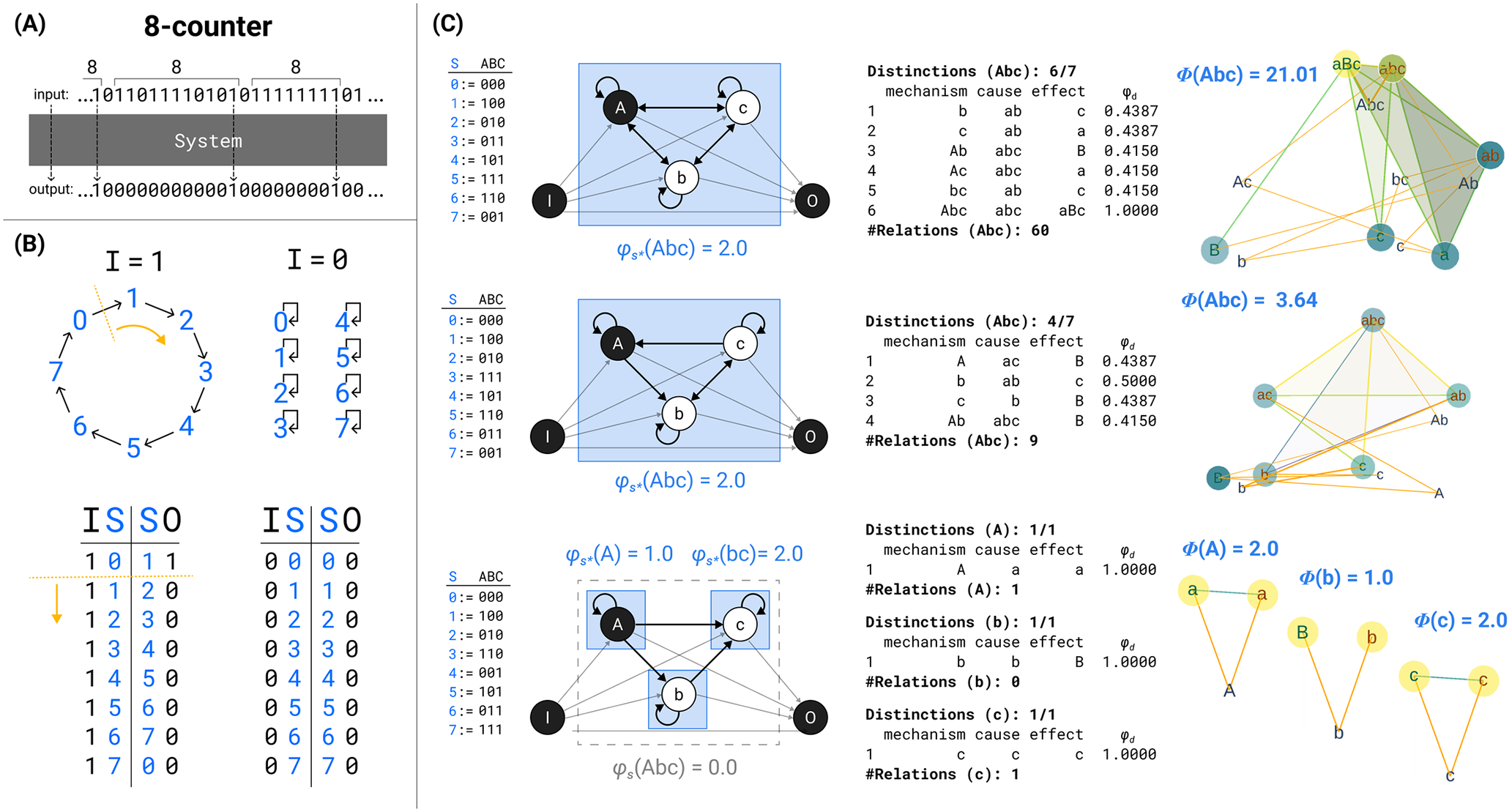}
\caption{\textbf{Functionally equivalent networks with different $\Phi$-structures.} (A) The input–output function realized by three different systems (shown in (C)): a count of eight instances of input I = 1 leads
to output O = 1. (B) The global state-transition diagram is also the same for the three systems: if I = 0,
the systems will remain in their current global state, labeled as 0-7; if I = 1, the systems will move one
state forward, cycling through their global states, and activate the output if S = 0. (C) Three systems
constituted of three binary units but differing in how the units are connected and interact. As a
consequence, the one-to-one mapping between the 3-bit binary states and the global state labels differ.
However, all three systems initially transition from 000 to 100 to 010. Analyzed in state 100, the first
system (top) turns out to be a single complex that specifies a $\Phi$-structure with six distinctions and many
relations, yielding a high value of $\Phi$. The second system (middle) is also a complex, with the same $\Phi$s value,
but it specifies a $\Phi$-structure with fewer distinctions and relations, yielding a lower value of $\Phi$. Finally, the
third system (bottom) is reducible ($\Phi$s = 0) and splits into two smaller complexes (entities) with minimal
$\Phi$-structures and low $\Phi$.}
\label{fig:iit}
\end{center}
\end{figure}

According to IIT, virtual implementability doesn’t imply consciousness and two substrates that are functionally equivalent do not have to be equivalent in their $\Phi$ structures. Interestingly, since a purely feed-forward system has $\Phi=0$ and any input-output function can be implemented by a feed-forward architecture, it follows that any behavior supported by a recurrent architecture can also be performed by an unconscious system. See Figure \ref{fig:iit}  from \cite{albantakis2023integrated}  for an example of three systems which implement the same function and have the same global dynamics but have a different causal structure and $\Phi$. 

\subsection*{\underline{10. Slicing problems and unity of experience (Level 3, Degree 3) }}
This challenge is from Nick Bostrom \cite{bostrom2006quantity}. The trick consists of considering a computer which can be separated in such a way that you end up with two separate computers running the same program (perhaps by gradually inserting an insulator on all the copper wires, or by inserting a randomization device). Percy and Gomez-Emilsson \cite{percy2022slicing} call this the slicing problem and illustrate it with a computer made of water logical units. According to computational functionalism, this should result in two identical conscious experiences, which for some might be counterintuitive. Bostrom argues that similar tricks can be performed to generate a fractional non-integer number of phenomenal experiences.\footnote{Even further, relating the slicing problem to the triviality problems above. If an insulator that divides a computer in two can duplicate a conscious experience, is the insulator even needed? Functionally, if the computer already implements the right mappings, doesn’t that same mapping already exist between the top half of the constituents and also between the electrons in the bottom halves of the wires? What is the relevance of parts and wholes of physical states that correspond to algorithmic states?}

Relatedly, Shiller \cite{shiller2024digital} discusses the temporal sequential intertwining of separate computational processes. If consciousness depends on computational runs, how does this interweaving affect phenomenal quality and subjectivity? Do the experiences merge? Are there slices of independently coherent experience experienced sequentially within a single subject? Is there an incoherent or blended experience? If one considers the implications of these questions, they can be cashed out as objections to digital consciousness \cite{gomez-emilsson2023boundary, simon2017hard}.

\subsection*{\underline{11. Topological Properties of Electromagnetic Fields (Level 3, Degree 3) }}
Electromagnetic fields (EMFs) in the brain involve not only action potentials in neurons, but also ion channels which modulate neural firing, ephaptic coupling (the firing of a neuron indirectly influencing other neurons \cite{han2020climbing}), and oscillatory waves at the population level, among others. 

While neuroscience has not traditionally assigned function to EMFs, this has increasingly changed in recent years. Earl Miller’s lab research, for example, has shown that the exact neurons maintaining a single memory vary from trial to trial (this is called representational drift), which implies representation has to happen at a higher level. This has led him to argue that the stability of working memory emerges at the level of electric fields and that different frequency bands carry different amounts of information \cite{pinotsis2023ephaptic}. Relatedly, a study found that neurons processing visual information fired asynchronously when an animal does not attend to the stimulus, but fired synchronously when the animal attends to, and is presumed to be conscious of, the stimulus \cite{kreiter1996stimulus}. This has led researchers to suggest that conscious experience depends on electromagnetic fields and have singled out several good properties of EMFs that would make them a good seat of consciousness.  For example, electromagnetic field propagation, unlike synaptic transition which is relatively slow, has the advantage of being nearly instantaneous \cite{mcfadden2023consciousness}. 

To be sure, oscillatory neural dynamics are not incompatible with computational functionalism, on the contrary, machine learning researchers have argued before that computational neural dynamics and trajectories (construed as attractor dynamics, which reflect oscillatory dynamics with harmonic modes) can explain the ineffability and richness of conscious states \cite{ji2024sources}. Conversely, EMF theories do not deny that much brain information processing can proceed via conventional neuron/synapse transmission \cite{mcfadden2020integrating}. 

Nevertheless, McFadden has explicitly predicted that AIs based on conventional computing will never be conscious. He argues it is the singular entity nature of EM fields – which provides a solution to the binding problem – where different aspects of an experience (color, objects, place) are bound into a single unit of experience : unity as harmony. This unity is also responsible for the fact that the conscious mind can only do one thing (or a few things, anyway) at a time. On a related note, others have argued that consciousness depends on topological properties of fields \cite{gomez-emilsson2023boundary} rather than on computational functional organization. 

\subsection*{\underline{12. Biological Complexity (Level 3, Degree 2)}}
Several scientists highlight the complexity of biological organization and the inextricable integration of sub-cellular, neural, and system levels which is non-existent in AI models. While these authors do not necessarily rule out the possibility of AI consciousness in principle, they often do suggest that current LLMs are not conscious and are not likely to be conscious soon. Here we present some of the most notorious:

Peter Godfrey Smith \cite{godfreysmith2016mind} points to the complexity of all the interactions and oscillatory waves that go way beyond the network-like neural connections of just the neurons. 

Similarly, Jaan Aru and colleagues \cite{aru2023feasibility} have pointed to the differences in architectural and functional organization between Large Language Models and biological brains, highlighting in the latter the higher degree of integration, both between external world and internal needs through “having skin in the game”, as well as across levels and substructures such as biological thalamic integration. 

Anil Seth \cite{seth2021being, ledoux2023consciousness} also believes that consciousness might be restricted to biological systems grounded in the biological imperative of physiological regulation, this imperative he argues, traverses all levels of biology and prohibits a distinction between mind and substrate that could be equivalent to that of software and hardware.

Perhaps more concretely, Rosa Cao \cite{cao2022multiple} points out that in the biological brain, functional properties are not nomologically independent of material properties. The functions of different processes of the brain (including the metabolic and informational processes) are not easily decomposed, but deeply intertwined and running simultaneously. Hence, she argues, for example, how Chalmers’ neural replacement scenario \cite{chalmers1996absent}, which is meant as a thought experiment arguing in favor of computational functionalism in which a replacement of neuron by neuron to silicon maintains the same functional organization, cannot really be realized. Instead, what would be required is a spatially varying and history-dependent sensitivity to a variety of biochemical impingements on the input side; and constant self-modification that changes the connectivity in both action potential generation and neurotransmitter release on the output side (see also Hinton's idea of 'mortal computation' \cite{hinton2022forward}). Fundamentally, these requirements provoke constraints that are not merely practical or technological; they are also nomological about the type of functions that brain structures implement.

\subsection*{\underline{13. Biology constrains consciousness through some fundamental property (Level 3, Degree 1) }}
Enactivists have argued that consciousness can only arise in life due to the tight interaction between organism and external environment. This results in systems that can create and maintain their own parts, terming this autopoiesis \cite{maturana1980autopoiesis}. According to Thompson, that self-producing property of biological life implies consciousness \cite{thompson2010mind}. Thus, consciousness lies not only on the brain but is also fundamentally sustained through the body and the environment. 

Although these theories are usually developed as positive accounts, rather than objections to computationalism, they are often presented as alternatives to computationalism. While they don’t necessarily rule out the possibility of digital consciousness, they do claim that consciousness (and cognition in living beings) does not arise in virtue of certain informational processes or computational organization, but in virtue of their capability of enacting the world where environments and organisms are codetermined and cotransformed \cite{varela1991embodied}. 

Similarly, adherents to the Free Energy Principle point to properties shared by self-organizing alive systems which might be incompatible with current artificial systems, examples include the thermodynamic cost and the causal flow associated with the computations \cite{wiese2024artificial, parr2022active}.

\subsection*{\underline{14. Quantum properties (Level 3, Degree 1)}} 
Perhaps consciousness is fundamentally related to the still not understood aspects of quantum reality. We mentioned how Penrose thought consciousness could transcend computability. Penrose suggests that this might ultimately be explainable because of the way in which wave function collapse contributes to the generation of consciousness. Penrose and Hameroff \cite{penrose1996orchestrated} have proposed that microtubules in neurons may be the site of the quantum-neural nexus, enabling forms of quantum computation. Others explore different connections between consciousness and the quantum. For example, there is a resurgence of interest in the idea that consciousness is necessary for wave function collapse (the converse of Penrose's view, in effect). See for example Chalmers and McQueen \cite{chalmers2022consciousness} and other entries in \cite{gao2022consciousness} for discussion of this and other possible quantum-consciousness connections (see for review Simon \cite{simon2024consciousness}]). A direct link between consciousness and the quantum realm suggests that there is more to consciousness than classical digital computation. How we taxonomize the constraint depends on how we fill in the details of the proposed link, but if we construe it, say, as the requirement that consciousness involve some quantum computation, then whether this classifies as Degree 1 or 3 hinges on how we delineate digital from non-digital forms of computation: is quantum computation digital in the relevant sense, or not? If yes, then the according worry is Degree 1, otherwise it is probably Degree 3.

\section*{Conclusion}
This work presented a taxonomy to situate the discussion around the possibility of digital consciousness, without arguing in favor or against. For those who believe digital consciousness is possible, it hopefully helped structure the debate and identify some conditions they need to answer. For those who are skeptical, this work perhaps helped to contextualize their reasons for skepticism and helps clarify what is the more precise nature of the existing arguments.

The taxonomy aimed to disentangle the challenges to computational functionalism from those to digital consciousness showing one can reject the former while accepting the latter (or vice versa). By organizing the constraints and objections into three different degrees of force and the three levels of Marr's hierarchy, one can disambiguate what otherwise can seem like monolithic positions (such as "enactivism" or “biological constraints”) into more precise claims. While objections at Level 2 tend to be more practical and compatible with the spirit of computational functionalism (if not necessarily with the existing definitions of algorithm), the other Levels more fundamentally challenge the framework. Either posing the importance of non-computable elements (Level 1) or invoking some organizational or fundamental intrinsic implementation orthogonal to computation (Level 3).

The nature of consciousness is far from settled. With the rapid progress of AI, we believe it will have important normative weight. This work aims to clarify and move forward the relevant underlying debate.

\pagebreak

\bibliographystyle{ieeetr}
\bibliography{main}  





\end{document}